\documentclass{Interspeech2024}
\interspeechcameraready 

\usepackage{multicol,multirow}
\usepackage{tabularx}
\usepackage{color, colortbl}
\usepackage[dvipsnames]{xcolor}
\definecolor{Gray}{gray}{0.9}
\usepackage{todonotes}

\title{A dual task learning approach to fine-tune a multilingual semantic speech encoder for Spoken Language Understanding\vspace{-0.3cm}}

\name[affiliation={1,2}]{Gaëlle}{Laperrière}
\name[affiliation={2}]{Sahar}{Ghannay}
\name[affiliation={1}]{Bassam}{Jabaian}
\name[affiliation={1}]{Yannick}{Estève}

\address{
  $^1$LIA - Avignon Université, France\\
  $^2$LISN - CNRS/Université Paris-Saclay, France}
\email{
    $^1$firstname.lastname@univ-avignon.fr, 
    $^2$firstname.lastname@lisn.upsaclay.fr
    \vspace{-0.3cm}}


\begin{document}

\maketitle

\begin{abstract} 
Self-Supervised Learning is vastly used to efficiently represent speech for Spoken Language Understanding, gradually replacing conventional approaches.
Meanwhile, textual SSL models are proposed to encode language-agnostic semantics.
SAMU-XLSR framework employed this semantic information to enrich multilingual speech representations.
A recent study investigated SAMU-XLSR in-domain semantic enrichment by specializing it on downstream transcriptions, leading to state-of-the-art results on a challenging SLU task.
This study's interest lies in the loss of multilingual performances and lack of specific-semantics training induced by such specialization in close languages without any SLU implication.
We also consider SAMU-XLSR's loss of initial cross-lingual abilities due to a separate SLU fine-tuning.
Therefore, this paper proposes a \textit{dual} task learning approach to improve SAMU-XLSR semantic enrichment 
while considering distant languages for multilingual and language portability experiments.
\end{abstract}


\vspace{-0.2cm}
\section{Introduction}
\label{sec:Introduction}

Spoken Language Understanding (SLU) implies extracting semantic information from speech signal \cite{tur2011spoken}. 
This Natural Language Processing task can be modeled as named entity recognition or slot filling in a Human-Machine dialogue context. 

This study focuses on a complex speech-to-concepts task by using end-to-end neural approaches as introduced by~\cite{ghannay2018end, haghani2018audio,serdyuk2018towards}. 
End-to-end approaches distinguish themselves from cascade approaches~\cite{liu2020mockingjay,liu2021tera,review2} by combining in a single model both Automatic Speech Recognition (ASR) and Natural Language Processing modules, resulting in less error propagation.

The main challenge of such an approach resides in the lack of paired speech recordings and semantic manual annotations. 
Many solutions have emerged, like transfer learning~\cite{bhosale2019end,Caubriere2019,huang2020leveraging}, data augmentation with speech synthesis~\cite{desot2020corpus,lugosch2020using} and the use of pre-trained self-supervised models.
Self-Supervised learning (SSL) enables the use of large amounts of unlabelled data.
This method, previously used for ASR~\cite{baevski2020wav2vec,devlin-etal-2019-bert}, was proven very useful for SLU cascade approaches~\cite{laperriere2021we} by enhancing both speech representations~\cite{liu2020mockingjay,liu2021tera,hsu2021hubert} and annotations~\cite{devlin-etal-2019-bert,feng2022language}.

However, combining text and speech SSL methods in a single end-to-end architecture is found very difficult due to the complexity of unifying the speech and textual representation spaces while optimizing a huge number of parameters.  
To exploit textual SSL models in end-to-end architectures,~\cite{wang2020large,chung2021splat} proposed to project ASR representations to a BERT model, while~\cite{huang2020leveraging,agrawal2022tie} carried out an intent classification by using sentence-level acoustic representations tied to a BERT SLU model. 
\cite{muller2021pursuit} proposed a similar multilingual approach.

Driven by this will to make use of speech SSL models in end-to-end architectures,~\cite{khurana2022samu} proposed SAMU-XLSR (Semantically-Aligned Multimodal Utterance-level Cross-Lingual Speech Representation).
This new method produces semantically enriched multilingual and multimodal speech representations by combining the multilingual speech encoder XLS-R~\cite{xlsr} to the Language Agnostic BERT Sentence Embedding generator LaBSE~\cite{feng2022language}.
Lately,~\cite{laperriere2022use} has demonstrated the pertinence of the model's semantically aware frame-level speech representations for a specific semantic extraction task on the challenging SLU MEDIA benchmark~\cite{bechet2019benchmarking}. 
More recently, we pursued this analysis~\cite{IS23} by demonstrating the pertinence of another semantic specialization of SAMU-XLSR representations on a small amount of transcribed data linked to the SLU task with language portability.

In this research work, our main objective is to leverage the availability of transcribed speech data from closely related domains in multiple languages to enhance the performance of an SLU model in languages with limited SLU annotations.
The use of the SAMU-XLSR model has proven to be particularly relevant in this case, but it seems perfectible. 
We assume that during the fine-tuning of the SAMU-XLSR model for an SLU task, it tends to forget its capacity to generate certain semantic abstractions at the utterance level prematurely. 
To limit this early forgetting, we propose a \textit{dual} task learning approach when fine-tuning the SAMU-XLSR to an SLU downstream task, further on referred to as ``\textit{dual} fine-tuning".

This approach is considered for the original French and Italian MEDIA benchmark used in~\cite{IS23} and the Tunisian TARIC-SLU dataset, freshly introduced by \cite{salima}.
By the use of two very low resource datasets, we investigate the multilinguality and cross-linguality gains of such \textit{dual} architecture compared to a more classic sequential approach. 
These experiments lead to state-of-the-art results with an end-to-end approach with all three datasets. 

\vspace{-0.2cm}
\section{SLU tasks in different languages}
\label{sec:Datasets}

In this study, we considered three datasets of different languages to conduct experiments on language portability and multilinguality: French (MEDIA), Italian (PortMEDIA) and the more distant Tunisian language (TARIC-SLU). 
All datasets were annotated for a complex semantic extraction task from speech, with close-domain labels. 
Our contributions focus on PortMEDIA and TARIC-SLU low-resource datasets, while MEDIA is mostly used to enhance their performances.

Table \ref{tab:Data} presents the audio duration and words distribution in each corpus for each of the three datasets described bellow.

\begin{table}[!ht]
    \begin{center}
        \vspace{-0.2cm}
        \caption{Hours and Words distribution in MEDIA, PortMEDIA and TARIC-SLU.}
        \vspace{-0.3cm}
        \begin{tabularx}{\columnwidth}{ c | c | c | c | c | c }
            \multicolumn{2}{c |}{} & lang & \textbf{Train} & \textbf{Valid} & \textbf{Test} \\
            \hline
            \multirow{3}{*}{Dur.}
            &\textbf{MEDIA} & 
             fr &  
             10h52m & 1h13m & 3h01m \\
            &\textbf{PortMEDIA} & it & 7h18m & 2h32m & 4h51m \\
            &\textbf{TARIC-SLU} & tu &   7h30m & 0h29m & 0h54m \\
            \hline   
            \multirow{3}{*}{Wrd}
            &\textbf{MEDIA} &  fr &  94.5k & 10.8k & 26.6k \\
            &\textbf{PortMEDIA} & it &  21.7k & 7.7k & 14.7k \\
            &\textbf{TARIC-SLU} & tu &  58.5k & 3.5k & 7.0k \\
            \hline  
        \end{tabularx}
        \vspace{-0.8cm}
        \label{tab:Data}
    \end{center}
\end{table}

\subsection{The French MEDIA dataset}
\label{sec:MEDIA}

The French MEDIA dataset \cite{BonneauMaynard2005}, part of ELRA's MEDIA Evaluation Package \footnote{\url{http://catalog.elra.info/en-us/repository/browse/ELRA-E0024/}} \footnote{ISLRN: 699-856-029-354-6}, is freely accessible for academic research, all recorded contributors having permitted its distribution. 

The MEDIA benchmark is composed of recorded phone calls for hotel booking, transcribed and annotated with semantic concepts. 
The recorded speech consists of Human-Machine dialogues collected with the Wizard-of-Oz method.
Only the user's turns are fully annotated and considered for this study. 

This dataset contains 1258 dialogues, from approximately 250 different French speakers.
Our following experiments were made with the \textsl{full} MEDIA version, containing much richer semantic annotations, for a total of 152 different concepts in comparison to the 76 concepts of the \textsl{relax} version.

\subsection{The Italian PortMEDIA dataset}
\label{sec:PortMEDIA}

The Italian PortMEDIA dataset \cite{lefevre-etal-2012-leveraging} is part of the same ELRA package \footnote{\url{http://www.elra.info/en/projects/archived-projects/port-media/}} as MEDIA.
Its annotations and transcriptions were collected in the same way as MEDIA's, for the exact same task. 

The dataset is composed of 604 dialogues from more than 150 different Italian speakers.
It uses a total of 139 semantic concepts \cite{bonneau-maynard-etal-2006-results} and is only available as a \textsl{full} version. 
Note that this is the smallest corpus used in this study, considering the number of words in the training set. 

\subsection{The Tunisian TARIC-SLU dataset}
\label{sec:TARIC}

The Tunisian TARIC-SLU dataset was sourced from the ASR TARIC one \cite{masmoudi}, dedicated to Tunisian Dialect Automatic Speech Recognition in the context of Human-Human dialogues.
It has then been semantically annotated for a semantic extraction task from speech, on top of intents of the speech segments which we do not study in this paper. 

The dataset's domain of annotation varies slightly from MEDIA's and PortMEDIA's.
Its speech recordings are conversations for train booking, of real conditions or acted dialogues.
$809$ out of $2'549$ concepts' values have code-switching, with an expected $80$\% of French words and $20$\% of English words.

A first version of TARIC-SLU, now updated and soon freely distributed, was proposed recently by \cite{salima}. 
TARIC-SLU is composed of more than 2,000 dialogues from 108 speakers, like its ASR version, and uses 60 different semantic concepts.

\subsection{Evaluation Metrics}
\label{sec:Metrics}

The semantic extraction task of this study aims to output the transcription of the speech signal and its annotation with semantic concepts as follows: \textsl{I \textnormal{{\textless}reservation\textgreater} would like to book \textnormal{\textgreater} \textnormal{{\textless}room-number\textgreater} one \textnormal{\textgreater} \textnormal{{\textless}room-type\textgreater} double room \textnormal{\textgreater}}. 

We consider two conventional metrics: the Concept Error Rate (CER) and the Concept Value Error Rate (CVER). 
Both metrics are computed like a Word Error Rate with the CER only accounting the hypothesis' semantic concepts while the CVER considers the correctness of each value (``double room'') and concept (``\textless room-type\textgreater'') pair as a single occurrence.

A relevant CER improvement should differ by $0.4\%$ considering MEDIA's test set, $0.7\%$ considering PortMEDIA's test set, and $1.0\%$ considering TARIC-SLU's test set. 
Please note a possible $0.3\%$ variation of Error Rates observed with 5 trainings of both SLU and \textit{dual} models presented in Section \ref{sec:SAMU}. 

\vspace{-0.2cm}
\section{SAMU-XLSR} 
\label{sec:SAMU}

In the domain of speech representation, self-supervised approaches have recently gained popularity over more conventional approaches such as filter-banks and MFCCs.
Their aptitude to leverage large masses of unlabelled speech has been proved very useful for many tasks. 
During the last few years, many models like wav2vec 2.0 \cite{baevski2020wav2vec}, HuBERT \cite{hsu2021hubert}, and WavLM \cite{chen2022wavlm} have proven their efficiency in ASR \cite{baevski2020wav2vec}, speaker verification \cite{chen2022large,chen22g_interspeech} and emotion recognition \cite{macary2021use,pepino21_interspeech}.
Following this surge for SSL, \cite{khurana2022samu} proposed the SAMU-XLSR approach for a translation task, aiming to capture the semantics in the speech signal by fine-tuning a pre-trained multilingual SSL model.

While higher-level semantics were already proven useful in speech-to-text mining and speech-to-text translation contexts, SAMU-XLSR's approach has been proven pertinent in other tasks such as semantic extraction from speech \cite{IS23}, where we specialized SAMU-XLSR and extracted its frame-level speech representations for an SLU task.

\begin{figure}[htb!]
\centering
    \vspace{-0.2cm}
    \includegraphics[scale=0.70]{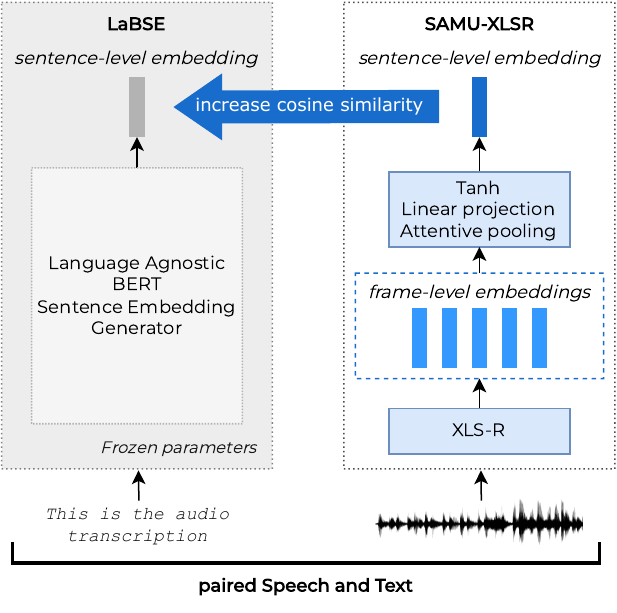} 
    \vspace{-0.3cm}
    \caption{Training and specialization process of SAMU-XLSR.}
    \vspace{-0.2cm}
    \label{fig:SAMU}
\end{figure}

Figure \ref{fig:SAMU} illustrates how SAMU-XLSR processes audio and text paired data. 
The pre-trained multilingual XLS-R \cite{xlsr} used in this approach was designed to generate speech representations for short 20 milliseconds speech frames. 
To make use of this model, SAMU-XLSR performs pooling and projection to create a single sentence-level representation. 
In parallel, LaBSE \cite{feng2022language} sentence-level textual representations are simply extracted. 
Both representations being on the same semantic space, SAMU-XLSR's is then being pulled towards LaBSE's with the help of a cosine similarity loss function. 
This means the parameters of all SAMU-XLSR's components are optimized to predict the textual representations generated by the frozen LaBSE model.  

In consideration to its semantic extraction performances, SAMU-XLSR has however been proven less apt to transcribe speech dialogues when specialized on multiple languages \cite{IS23}.

\subsection{SLU fine-tuning} 
\label{sec:SLU}

\begin{figure}[htb!]
\centering
    \vspace{-0.2cm}
    \includegraphics[scale=0.70]{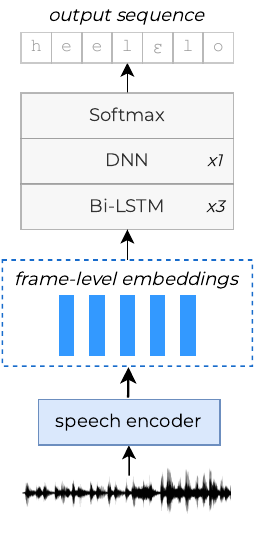} 
    \vspace{-0.5cm}
    \caption{Fine-tuning process of SAMU-XLSR for an SLU task.}
    \vspace{-0.3cm}
    \label{fig:SLU}
\end{figure}

Fine-tuning SAMU-XLSR speech encoder on a specific SLU task initially required an SLU model, such as the one used in \cite{IS23} and presented in Figure \ref{fig:SLU}.
This end-to-end model was used after SAMU-XLSR specialization on the downstream task, while this study presents an architecture to do both specialization and fine-tuning steps in one single training.
We compared our \textit{dual} approach to the previous architecture and tried out a second fine-tuning on TARIC-SLU in section \ref{sec:Distant}.
This SLU model was also necessary to alleviate the lack of baseline on the TARIC-SLU dataset. 

The architecture consists of a fine-tuned speech encoder (specialized or dually fine-tuned SAMU-XLSR) generating frame-level speech representations of the input signal. 
These are given to 3 bi-LSTM layers of 1024 neurons, contextualizing the audio segments. 
A DNN layer of the same dimension, activated by LeakyReLU, feeds its outputs to a softmax function.
We optimize a greedy loss with an Adam optimizer of $learning\_rate=0.0001$ for both speech encoder and Bi-LSTMs, and Adadelta with $learning\_rate=1.0$ for the DNN.

After trying out other upper modules with different dimensions to decode the speech representations, we decided to keep the bi-LSTMs, proven more efficient than DNNs for this task.

\subsection{Dual task learning} 
\label{sec:Dual}

SAMU-XLSR approach allows better capture of semantics directly from the speech signal for multiple languages \cite{laperriere2022use,khurana2022samu}.

This study aims to improve the specialization and fine-tuning processes on multilinguality and specific semantics with the help of distant languages and SLU modality. 
By merging the SLU fine-tuning and specialization, we expect to prevent suppressing SAMU-XLSR capacities on capturing internal utterance-level semantic abstractions during the fine-tuning of the SLU downstream task.
To address this phenomenon of forgetting, we propose a \textit{dual} task learning approach, wherein the SAMU-XSLR model is fine-tuned to:

\begin{itemize}
\item Generate utterance-level embeddings, similar to its initial pre-training, but only for utterances relevant to the SLU downstream task, potentially across multiple languages.
\item Perform the SLU downstream task in the targeted language.
\end{itemize}


\begin{figure*}[htb!]
\centering
    \vspace{-0.5cm}
    \includegraphics[scale=0.70]{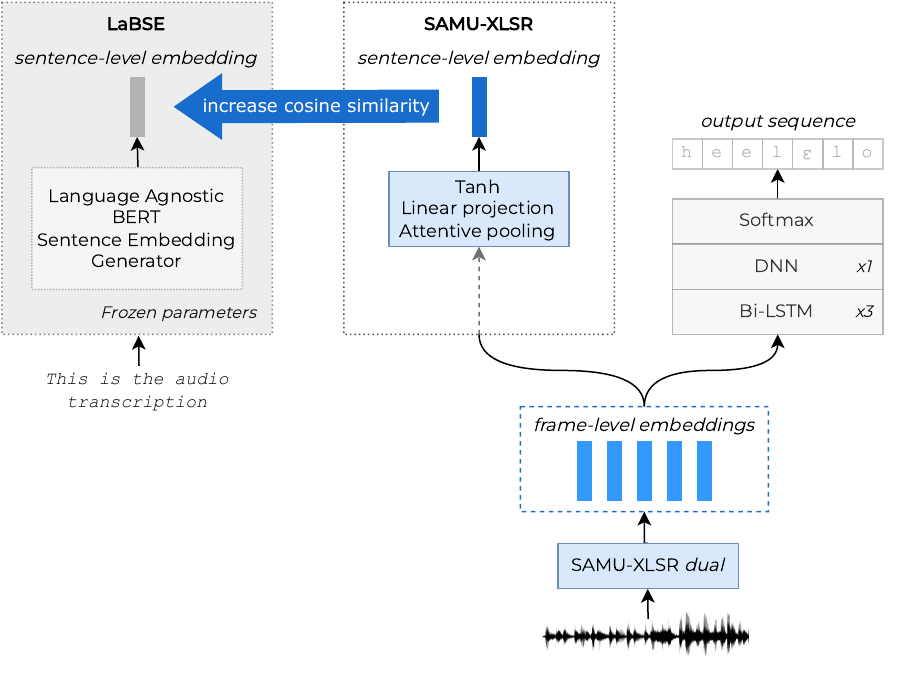} 
    \vspace{-0.5cm}
    \caption{Training process of the SLU and SAMU-XLSR modules combined in a \textit{dual} architecture.}
    \label{fig:Dual}
    \vspace{-0.5cm}
\end{figure*}

Figure \ref{fig:Dual} schematizes the \textit{dual} architecture with the speech encoder connection to both SAMU-XLSR and SLU upper modules. 
For comparison purposes, both architectures presented before have been reused in this single model.

The only hyper parameter search done for this architecture concerns the loss distribution between both modules. 
We considered this distribution as follows: $loss = loss(SAMU\-XLSR) + \lambda \ loss(SLU)$, with multiple $\lambda$ tested in the interval $[0;20]$, especially between $0$ and $1$. 

\vspace{-0.2cm}
\section{Experimental results}
\label{sec:Results}

This section presents experimental results of the \textit{dual} approach (SAMU-XLSR$_{dual \ L}$ trained on languages $L$) by comparing them to a simple specialization (SAMU-XLSR$_{L}$) followed by a necessary SLU training (SLU$_{L}$). 
Datasets were not shuffled during training, having previously experimented on multiple shuffling methods without evident benefits from them. 
Note that only SAMU-XLSR was considered for this study, being proven the current most pertinent speech encoder for these SLU tasks, with presently the best results on MEDIA and PortMEDIA \cite{IS23}.

\subsection{Task-oriented semantic enrichment} 
\label{sec:Monolingue}

As we saw a frank improvement of CER in giving more importance to the SLU loss than SAMU-XLSR's loss for MEDIA experiments, we focused on optimizing the speech encoder mostly on the SLU task while specializing it more slightly on the SAMU-XLSR task. 
However, contrary to experiments with a $\lambda$ in a $[0;1]$ interval, no linear improvements were observed when this coefficient was defined between $1$ and $20$.
Therefore, all results were obtained by running experiments with multiple $\lambda$ in the $[1;20]$ interval and keeping the best system considering the development set.  
Results of monolingual \textit{dual} fine-tuning on each dataset are presented for the test set in Table \ref{tab:Monolingue}. 

\begin{table}[!ht]
    \begin{center}
    \vspace{-0.2cm}
    \caption{Experimental results on \textit{dual} approach for a monolingual fine-tuning, compared to a classic specialization followed by an SLU fine-tuning.}
    \vspace{-0.4cm}
    \resizebox{\columnwidth}{!}{
        \begin{tabular}{ l | l | c | c |}
            \multicolumn{2}{c |}{} & CER & CVER \\
            \hline
            \multirow{3}{*}{fr}
            & \cellcolor{Gray}SAMU-XLSR $+$ SLU $_{FR}$ & $18.7$ & $29.4$ \\
            & \cellcolor{Gray}SAMU-XLSR $_{FR}$ $+$ SLU $_{FR}$ \cite{IS23} & $18.5$ & $29.5$ \\
            & SAMU-XLSR $_{FR \ dual}$ & \textbf{18.3} & \textbf{27.8} \\
            \hline
            \multirow{3}{*}{it}
            & \cellcolor{Gray}SAMU-XLSR $+$ SLU $_{IT}$ \cite{IS23} & \textbf{26.6} & \textbf{39.2} \\
            & \cellcolor{Gray}SAMU-XLSR $_{IT}$ $+$ SLU $_{IT}$ \cite{IS23} & $26.8$ & $39.5$ \\
            & SAMU-XLSR $_{IT \ dual}$ & $26.8$ & $39.4$ \\
            \hline
            \multirow{4}{*}{tu}
            & \cellcolor{Gray}SAMU-XLSR $+$ SLU $_{TU}$ & $30.7$ & $47.4$ \\
            & \cellcolor{Gray}SAMU-XLSR $_{TU}$ $+$ SLU $_{TU}$ & $30.3$ & \textbf{45.2} \\
            & SAMU-XLSR $_{TU \ dual}$ & $32.4$ & $48.3$ \\
            & SAMU-XLSR $_{TU \ dual}$ $+$ SLU $_{TU}$ & \textbf{29.9} & $46.8$ \\
            \hline  
        \end{tabular}
    }
    \label{tab:Monolingue}
    \vspace{-0.5cm}
    \end{center}
\end{table}

The \textit{dual} task learning leads to relatively baseline equivalent CERs for monolingual experiments. 
The main advantage of this method resides in the number of parameters learned on a single v100-32G GPU during 100 epochs for each training. 
SAMU-XLSR specialization optimizes $316.2$M parameters, in addition to $387.8$M for SLU fine-tuning, while the \textit{dual} fine-tuning only optimizes $385.6$M parameters.
Note that the Tunisian task requires a supplementary SLU fine-tuning to yield pertinent results.
This can be explained by the lack of Tunisian data in SAMU-XLSR's original training, leading to its seemingly less noteworthy fine-tuning. 
Indeed, where training on French and Italian means fine-tuning SAMU-XLSR on these languages, training on Tunisian means a first-time SAMU-XLSR training on this dialect, done at the same time as an SLU training. 
Multilingual \textit{dual} fine-tuning will prove itself efficient on never-seen data such as Tunisian in the following language portability experiments. 

\subsection{Language portability} 
\label{sec:Multilingue}

State-of-the-art results for all three datasets were obtained with language portability.
We can distinguish experiments on close languages like French and Italian, and ones on distant languages with the use of these two to improve performances on the Tunisian TARIC-SLU task. 

\vspace{-0.2cm}
\subsubsection{Close languages}
\label{sec:Close}

Close-language specialization was conducted in \cite{IS23} to make use of in-domain data. 
This paper studies the use of the \textit{dual} architecture in a multilingual context.
Table \ref{tab:Close} shows experimental results with French and Italian mixed fine-tunings. 

\begin{table}[!ht]
    \begin{center}
    \vspace{-0.2cm}
    \caption{Experimental results on \textit{dual} approach with close-language portability.}
    \vspace{-0.3cm}
    \resizebox{\columnwidth}{!}{
        \begin{tabular}{ l | l | c | c |}
            \multicolumn{2}{c |}{} & CER & CVER \\
            \hline
            \multirow{2}{*}{fr}
            & \cellcolor{Gray}SAMU-XLSR $_{FR \oplus IT}$ $+$ SLU $_{FR}$ & $18.6$ & $29.1$ \\
            & SAMU-XLSR $_{FR \oplus IT \ dual}$ & \textcolor{NavyBlue}{\textbf{17.9}} & \textbf{28.2} \\
            \hline
            \multirow{2}{*}{it}
            & \cellcolor{Gray}SAMU-XLSR $_{FR \oplus IT}$ $+$ SLU $_{FR \rightarrow IT}$ \cite{IS23} & $25.1$ & \textbf{38.1} \\
            & SAMU-XLSR $_{FR \oplus IT \ dual}$ & \textcolor{NavyBlue}{\textbf{24.1}} & $39.0$ \\
            \hline  
        \end{tabular}
    }
    \label{tab:Close}
    \vspace{-0.5cm}
    \end{center}
\end{table}

These experiments lead to a pertinent improvement of state-of-the-art CER results on MEDIA full version task with a score of \textbf{17.9}$\%$ and on PortMEDIA with a score of \textbf{24.1}$\%$. 

\vspace{-0.2cm}
\subsubsection{Distant languages}
\label{sec:Distant}

The main limitation experimented in \cite{IS23} was the lack of distant-language data to use during specializations. 
This paper studies the Tunisian language on which SAMU-XLSR has never been trained. 
Table \ref{tab:Distant} presents experimental results for this low-resource dataset with the use of French or Italian data. 

\begin{table}[!ht]
    \begin{center}
    \caption{Experimental results on \textit{dual} approach with distant-language portability for the Tunisian TARIC-SLU dataset.}
    \vspace{-0.3cm}
    \resizebox{\columnwidth}{!}{
        \begin{tabular}{ l | c | c |}
            & CER & CVER \\
            \hline
            \cellcolor{Gray}SAMU-XLSR $_{FR \oplus IT \oplus TU}$ $+$ SLU $_{TU}$ & $31.9$ & $47.0$ \\
            SAMU-XLSR $_{FR \ dual}$ $+$ SLU $_{TU}$ & $30.3$ & $46.4$ \\
            SAMU-XLSR $_{IT \ dual}$ $+$ SLU $_{TU}$ & $30.8$ & $46.4$ \\
            SAMU-XLSR $_{FR \oplus IT \ dual}$ $+$ SLU $_{TU}$ & $30.4$ & $48.4$ \\
            SAMU-XLSR $_{FR \oplus IT \oplus TU \ dual}$ $+$ SLU $_{TU}$ & \textcolor{NavyBlue}{\textbf{29.1}} & \textbf{46.2} \\
            \hline  
        \end{tabular}
    }
    \label{tab:Distant}
    \vspace{-1cm}
    \end{center}
\end{table}

During all \textit{dual} fine-tunings on Tunisian data, models hardly succeeded to individually focus on Tunisian while learning to represent semantics in French and Italian.
However, both languages were highly beneficial to TARIC-SLU processing with another SLU fine-tuning on Tunisian, as shown by the state-of-the-art \textbf{29.1}$\%$ CER obtained with a \textit{dual} fine-tuning on all three datasets, in comparison to the $29.9\%$ CER with the Tunisian-only \textit{dual} fine-tuning of Table \ref{tab:Monolingue}.


\vspace{-0.2cm}
\section{Conclusion}
\label{sec:Conclusion}

This paper investigates a new approach to improve the semantic enhancement of SAMU-XLSR speech representations for a complex SLU task. 
The \textit{dual} architecture proposed combines both SAMU-XLSR's original training process and an SLU module for semantic extraction, implying almost two times less parameters to optimize. 
This new architecture aims to prevent an expected loss of multilinguality and cross-linguality of SAMU-XLSR first pre-training, while filling in the lack of specific semantics focus during a classic specialization. 
Contrarily to the method proposed in \cite{IS23}, this study experiments on distant language's portability such as French and Italian for a targeted low-resource Tunisian task. 
This induced another challenge through the lack of Tunisian data in SAMU-XLSR original pre-training. 
However, the \textit{dual} fine-tuning, followed by a necessary SLU fine-tuning on this language leads to state-of-the-art $29.1\%$ CER. 
At the same time, with a single \textit{dual} training, MEDIA and PortMEDIA datasets outperformed state-of-the-art results with, respectively, $17.9\%$ and $24.1 \%$ CER scores. 
One limitation of this paper could be the omission of out-of-domain or less specialized data during the \textit{dual} fine-tuning, which was not experimented due to our focus on improving the already proven efficient SAMU-XLSR specialization and SLU fine-tuning on MEDIA and PortMEDIA. 
However, the \textit{dual} task learning approach experimented on Tunisian SLU opens promising perspectives for other never-seen low-resource languages during the pre-training of SSL models, with a way to expand or specialize their multilinguality disposition.


\pagebreak
\section{Acknowledgements}
\label{sec:Acknowledgements}

This work was performed using HPC resources from GENCI/IDRIS (grant 2022 AD011012565) and received funding from the EU H2020 SELMA (grant No 957017), ESPERANTO (grant No 101007666) and PSPC AIDA: 2019-PSPC-09 (funded by BPI-France) projects.

\bibliographystyle{IEEEtran}
\bibliography{mybib}

\end{document}